\documentclass[sigconf]{acmart}

\usepackage{epsfig}
\usepackage{graphicx}
\usepackage{amsmath}
\usepackage{amssymb}
\usepackage{CJKutf8}

\usepackage{url}
\usepackage{tabularx}
\usepackage{multirow}
\usepackage{subfigure}
\usepackage{amsmath,dsfont}
\usepackage{array}

\usepackage{algorithm} 
\usepackage{algorithmic} 

\usepackage{pifont}

\usepackage{verbatim}

\usepackage{colortbl}
\definecolor{mygray}{gray}{.75}

\usepackage{booktabs}

\usepackage{xcolor}

\usepackage{hyperref}

\newcommand{\blue}[1]{\textcolor{blue}{#1}}

\usepackage[normalem]{ulem}

\newcommand{\argmin}{\operatornamewithlimits{argmin}}

\newcommand{\eg}{\emph{e.g.,}~}
\newcommand{\etal}{\emph{et al.}~}
\newcommand{\ie}{\emph{i.e.,}~}

\newcommand{\specialcell}[2][l]{%
  \begin{tabular}[#1]{@{}l@{}}#2\end{tabular}}

\newcommand{\PreserveBackslash}[1]{\let\temp=\\#1\let\\=\temp}
\newcolumntype{C}[1]{>{\PreserveBackslash\centering}p{#1}}
\newcolumntype{R}[1]{>{\PreserveBackslash\raggedleft}p{#1}}
\newcolumntype{L}[1]{>{\PreserveBackslash\raggedright}p{#1}}

\newfont{\mycrnotice}{ptmr8t at 7pt}
\newfont{\myconfname}{ptmri8t at 7pt}
\begin{document}
\begin{CJK*}{UTF8}{gbsn} 
\newcommand{\graybg}[1]{\colorbox{gray!15}{#1}}

\copyrightyear{2017}
\acmYear{2017}
\setcopyright{acmlicensed} 
\acmConference{MM '17}{}{October 23--27, 2017, Mountain View, CA, USA.}
\acmPrice{15.00} 
\acmDOI{https://doi.org/10.1145/3123266.3123366}
\acmISBN{ISBN 978-1-4503-4906-2/17/10}

\title{Fluency-Guided Cross-Lingual Image Captioning}


\author{Weiyu Lan}
\affiliation{\institution{MMC Lab, School of Information\\Renmin University of China}}

\author{Xirong Li}
\authornote{Corresponding author (xirong@ruc.edu.cn).}
\affiliation{\institution{Key Lab of DEKE\\Renmin University of China}}

\author{Jianfeng Dong}
\affiliation{\institution{College of Computer Science and Technology, Zhejiang University}}


\begin{abstract}
Image captioning has so far been explored mostly in English, as most available datasets are in this language. However, the application of image captioning should not be restricted by language.
Only few studies have been conducted for image captioning in a cross-lingual setting.
Different from these works that manually build a dataset for a target language,
we aim to learn a cross-lingual captioning model fully from machine-translated sentences.
To conquer the lack of fluency in the translated sentences, we propose in this paper a fluency-guided learning framework.
The framework comprises a module to automatically estimate the fluency of the sentences and another module to utilize the estimated fluency scores to effectively train an image captioning model for the target language.
As experiments on two bilingual (English-Chinese) datasets show, our approach improves both fluency and relevance of the generated captions in Chinese, but without using any manually written sentences from the target language. 
\end{abstract}



\maketitle



\section{Introduction} \label{sec:intro}

Given a picture, human can give a concise description in the form of a well-organized sentence, identifying salient objects in the image and their relationship with the surrounding. But for computers, image captioning is a challenging task. Not only does the computer need to capture concise concepts in the picture, but it also has to learn a language model that generates proper sentences. 
Aided by advances in training deep neural networks and large datasets that associate images with text, recent works have significantly improved the quality of caption generation~\cite{google-show-tell,cvpr2015-neuraltalk,iclr15-ma-mrnn,wang2016image,li2016image}. 

\begin{figure}[tb!]
\centering
\noindent\includegraphics[width=1.01\columnwidth]{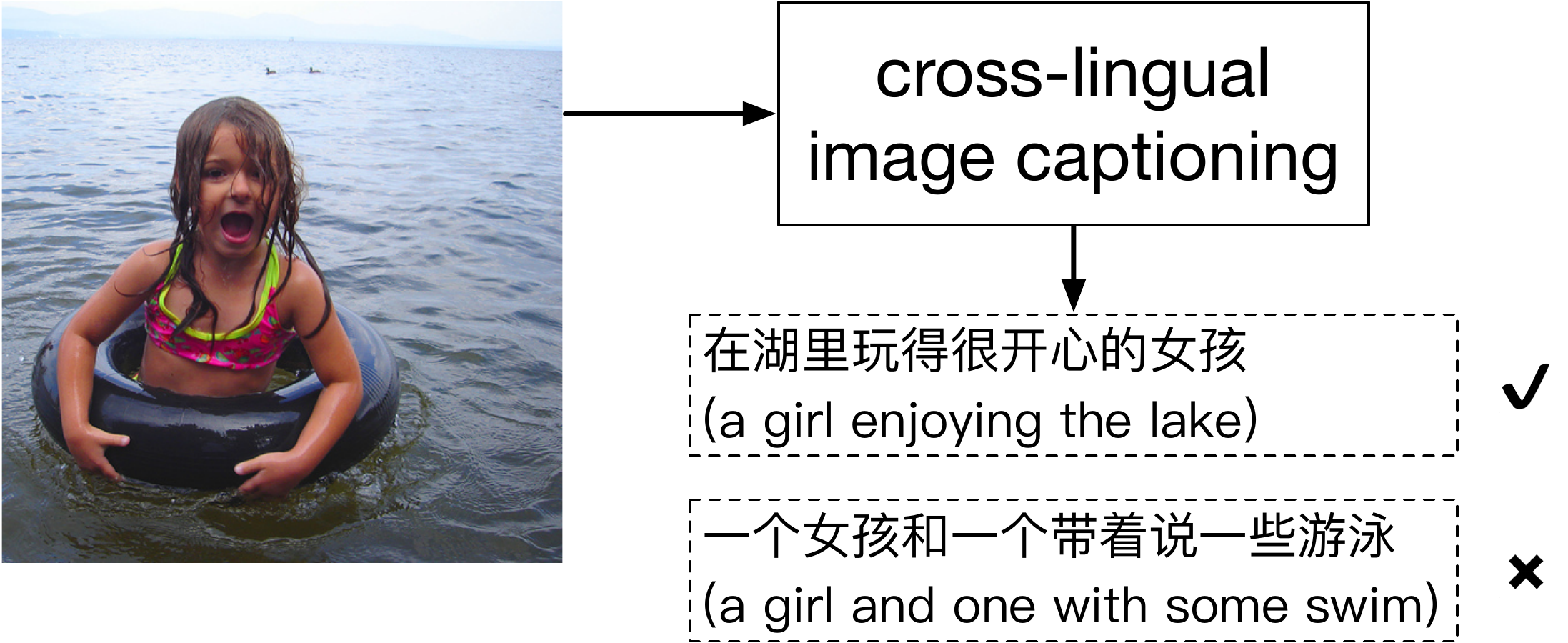}
\caption{This paper contributes to cross-lingual image captioning, aiming to generate relevant and fluent captions in a target language but without the need of any manually written image descriptions in that language. Manual translation of the Chinese sentences is provided in the parenthesis for non-Chinese readers.}
\label{fig:conceptual}
\end{figure}

With few exceptions, the task of image caption generation has so far been explored only in English since most available datasets are in this language. The application of image captioning, however, should not be restricted by language. The study of cross-lingual image captioning is essential for a large population on the planet who cannot speak English.
In this paper, we study cross-lingual image captioning that aims to generate captions in another language, as exemplified in Figure \ref{fig:conceptual}.
We target at Chinese, which is the most spoken language on the earth yet undeveloped in the image captioning research.

Only few studies have been conducted for image captioning in a cross-lingual setting \cite{elliott2015multilingual,miyazaki-cross,icmr2016}. 
They tackle this problem by constructing a new dataset in the target language.
Such an approach is constrained by the availability of manual annotation, and thus difficult to scale up and cover other languages. 
Instead of building a large dataset in a new language manually, we target at learning from machine-translated text.



While the use of web-scale data has substantially improved machine translation quality \cite{bahdanau2014neural, google-translation, baidu-translation}, we observe that the fluency of machine-translated Chinese sentences is often unsatisfactory. Fluency here means ``the extent to which each sentence reads naturally''~\cite{hovy2002principles}.
For instance, the sentence `A couple sit on the grass with a baby and stroller' is translated to `一对夫妇坐在婴儿推车的草' by Baidu translation, which is among the best English-to-Chinese translation systems. 
The keywords in the sentence are basically correctly translated, but the inappropriate conjunction of sentence elements makes the translated sentence not fluent. It tends to get even worse as English sentences becomes longer.
Due to the lack of fluency, directly learning a cross-lingual image captioning model from machine-translated text is problematic. For the same reason, directly translating the output of an English captioning model is questionable also. Moreover, the generated English captions are not always relevant to the image content, and the irrelevant part can be exaggerated via translation.


To conquer the obstacle of exploiting machine-translated text, we propose in this paper fluency-guided learning. Instead of revising the translated sentences to make them more fluent, which remains open in machine translation, we introduce a neural classifier to automatically estimate the fluency of these sentences. This provides an effective means to measure the importance of the sentences for training. For instance, sentences with lower fluency scores tend to be excluded from training or have a reduced effect on the captioning model. We make the intuition concrete by introducing three fluency-guided learning strategies. Automated and human evaluations on two datasets show the viability of the proposed framework. Code and data are available at \blue{\url{https://github.com/weiyuk/fluent-cap}}.

%
The rest of the paper is organized as follows. 
We review recent progress on image captioning in Section \ref{sec:related}.
We then propose our strategies in Section \ref{sec:approach}.
A quantitative evaluation is given in Section \ref{sec:eval},
with major findings reported in Section \ref{sec:conc}.

\section{Progress on Image Captioning} \label{sec:related}

\textbf{Monolingual image captioning}.
Our approach is developed on the top of monolingual image captioning. So we will first review recent progress in this direction. 
Three leading approaches have been explored~\cite{caption-survey}. 
The first formulates image captioning as a retrieval problem. Hodosh \etal \cite{flickr8k} propose to exploit similarity in the visual space to transfer candidate training descriptions to a query image.
Some other works, \eg \cite{nips13devise, jia2011learning, socher2014grounded} similarly rank existing descriptions but in a common multimodal space for the visual and textual data.
Following the progress in object detecting, detection based approaches \cite{farhadi2010every,cvpr2015-mscaption} generate descriptions using templates or grammar rules or language models based on the detected attributes of the objects in the image. 
Farhadi \etal \cite{farhadi2010every}, for instance, fill a fixed template by an inferred triplet of scene elements. More recently, Fang \etal \cite{cvpr2015-mscaption} uses a deep convolutional neural network (CNN) to predict a number of words that are likely to be present in a caption and generates description by a maximum-entropy language model. 
This approach constrains the diversity of generated descriptions as it relies on a predefined set of words or semantic concepts of objects, attributes and actions.

The recent dominant line in image captioning, inspired by the success of deep learning in image classification and sequence generation, is to apply deep neural networks which typically contain a CNN and an RNN to automatically generate new captions for images. 
In \cite{google-show-tell,cvpr2015-neuraltalk,nips15mmqa} , a CNN pretrained on the ImageNet classification task is used to encode an image, and a Recurrent Neural Network (RNN) is then used to decode the visual representation, outputting a sequence of words as the caption.
Xu \etal \cite{xu2015show} introduced an attention mechanism that incorporates visual context during sentence generation.
More recently, using scene information~\cite{li2016image} and high-level concepts / attributes as visual representation \cite{wu2016value} or as an external input for RNN \cite{yao2016boosting} is shown to obtain encouraging improvements over a standard CNN-RNN image captioning model.
Some new  architectures are continuous developed. For instance, Wang \etal \cite{wang2016image} propose a deeper bidirectional variant of Long Short Term Memory (LSTM) to take both history and future context into account in image captioning. A concept and syntax transition network \cite{karayil2016generating} is presented to deal with large real-world captioning datasets such as YFCC100M \cite{thomee2016yfcc100m}. 
Furthermore, in \cite{rennie2016self}, reinforcement learning is also utilized to train the CNN-RNN based model directly on test metrics of the captioning task, showing significant gains in performance.
We take a direction orthogonal to these works, aiming to exploit an existing model in the new cross-lingual context.
Hence, our work naturally benefits from the continuous progress in monolingual image captioning.


\textbf{Cross-lingual image captioning}.
Comparing to the large amount of interests in studying how to generate English captions, few studies have been conducted on cross-lingual image captioning. Elliott \etal \cite{elliott2015multilingual} address this topic as a translation problem, generating a description in the target language for a given image with a strong assumption that source-language descriptions are already provided for the image. 
To train a Japanese captioning model, Miyazaki and Shimizu \cite{miyazaki-cross}  use crowd sourcing to collect Japanese descriptions of the MSCOCO training set \cite{mscoco}. 
Different from the above works that require image descriptions manually written in the target language, our approach trains a cross-lingual image captioning model on machine-translated text. 
Li \etal \cite{icmr2016} have made a first attempt in this direction.
However, they use the translated text as it is, directly training a Chinese captioning model using machine-translated sentences from the Flickr8k dataset \cite{flickr8k}. As such, their model tends to generate Chinese captions with
ill­formed structures and thus bad user experience as exemplified in Fig. \ref{fig:conceptual}. The fluency problem is completely untouched in their model training and evaluation.



\begin{figure*}[tb!]
\centering
\noindent\includegraphics[width=2.05\columnwidth]{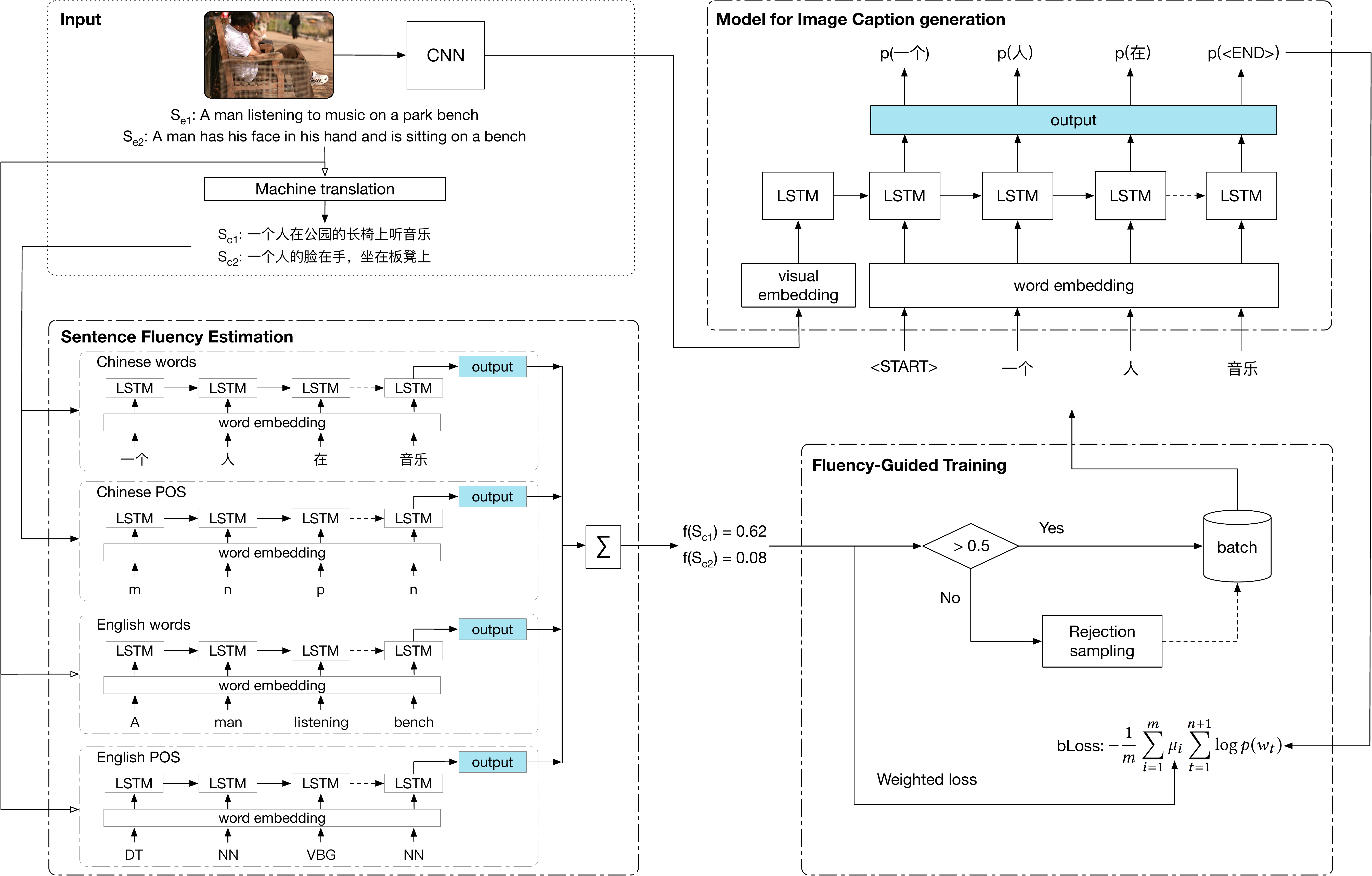}
\caption{The proposed fluency-guided learning framework for cross-lingual image captioning. 
Given English sentences $\{S_{e}\}$ describing a given set of training images, 
we first employ machine translation to generate Chinese sentences $\{S_c\}$.
A four-way LSTM based classifier assigns $f(S_c)$, a probabilistic estimate of each translated sentence being fluent.
The fluency scores are exploited by distinct strategies, \eg rejection sampling or weighted loss
to guide the learning process to emphasize training examples with higher fluency scores. As such, without the need of using any \textit{manually written} Chinese sentences in the training stage, the resultant image captioning model is capable of generating well-formed Chinese captions for novel images.}
\label{fig:framework}
\end{figure*}

\section{Our Approach} \label{sec:approach}



Our goal is to build an image captioning model for a target language, but without the need of any \textit{manually written} captions in that language for training. This is achieved by a novel cross-lingual use of training corpus from a source language.
Because public datasets for image captioning are in English \cite{flickr8k,flickr30k,mscoco} while Chinese is the most spoken language in the world, we consider English-to-Chinese as the cross-lingual setting.
Let $\{S_{e}\}$ be English sentences describing a given set of training images. 
Performing machine translation on these sentences allows us to automatically obtain their Chinese counterparts $\{S_c\}$.
As we have noted, the main challenge in learning an image captioning model from $\{S_c\}$ is that many of the machine translated sentences lack fluency. 
To conquer the challenge, the fluency of the training sentences needs to be taken into account.
To this end, we proposed a fluency-guided learning framework, as illustrated in Fig. \ref{fig:framework}.
We introduce Sentence Fluency Estimation as an automated measure of the fluency of each translated sentence. 
We then exploit the estimated fluency to guide the learning process to emphasize better translated sentences.
Individual components of the proposed framework are detailed as follows.

\subsection{Sentence Fluency Estimation}

It is worth noting that we do not intend to revise $\{S_c\}$ to make them more fluent, as this remains an open problem in machine translation \cite{stymne2013statistical,chang2009improving}.
Rather, we aim to automatically measure their fluency so that we might discard sentences that are deemed to be not fluent or minimize their effect during the training process. 
As a given sentence can be either fluent or not fluent, we approach the problem of sentence fluency estimation by binary classification.

In order to construct a classifier for sentence fluency estimation, we need to encode sentences of varied length into fixed-size feature vectors, and build a specific classifier on the top of the features. LSTM \cite{lstm}, for its capability of modeling long-term word dependency in natural language text, has been used to learn a meaningful and compact representation for a given sentence \cite{tacl2015-kiros, emnlp2016-mcb}. 
We therefore develop an LSTM based classifier, using the LSTM module for sentence encoding followed by a fully connected layer for classification.
Suppose we have access to a set of labeled sentences $\mathcal{D}=\{S_c, y\}$ where $y=1$ indicates the translated sentence is fluent and $y=0$ otherwise. 
Unlike western languages, many east Asian languages including Chinese are written without explicit word delimiters. Therefore, word segmentation is performed to tokenize a given sentence to a sequence of Chinese words.
We employ BOSON~\cite{boson}, a cloud based platform providing rich Chinese natural language processing service. 
Given $S_c$ as a sequence of $n$ words $(w_1, w_2, \ldots, w_n)$,
we feed the embedding vector of each word into the LSTM module sequentially, using the hidden state vector at the last time step as the feature vector $h(S_c)$. The vector then goes through the classification module, yielding two outputs $f(S_c)$ and $\hat{f}(S_c)$ indicating the probability of the sentence being fluent and not fluent, respectively. More formally, we have 
\begin{equation}
(f(S_c), \hat{f}(S_c)) = \mbox{softmax}(W \cdot h(S_c) + b),
\end{equation}
where $W$ is affine transformation matrix and $b$ is a bias term. 
We optimize the encoding module and the classification module jointly, representing all the parameters by $\Theta = [W_e, W, b, \phi]$, where $W_e$ is the word embedding matrix, and $\phi$ parameterizes affine transformations inside LSTM.
We train the classifier by minimizing the cross-entropy loss:
\begin{equation}\label{eq:obj}
\argmin_{\Theta} \sum_{(S_c,y)\in \mathcal{D}} -  \left( y \cdot \log(f(S_c)) + (1-y) \cdot \log(\hat{f}(S_c))\right). 
\end{equation}

As the Chinese sentences are generated by machine translation, a not fluent $S_c$ means the corresponding $S_e$ is difficult to be translated. 
Hence, the original English sentences might be another clue for sentence fluency estimation.
Moreover, as part of speech (POS) tags of a Chinese / English sentence reflects to some extent grammatical structures of the sentence,
they might be helpful for fluency estimation as well. 
In that regard, we train three more LSTM based classifiers,
denoted as $f(S_e)$, $f(S_{c,pos})$, and $f(S_{e,pos})$,
which respectively takes a sequence of English words, a sequence of Chinese POS tags and a sequence of English POS tags as input.
As a consequence, we obtain a four-way LSTM based classifier, which predicts the fluency of a translated sentence by combining the prediction of the four individual classifiers, \ie
\begin{equation}
f(S_c) \leftarrow \frac{1}{4} (f(S_c) + f(S_{c,pos}) + f(S_e) + f(S_{e,pos})).
\end{equation}
A translated Chinese sentence is classified as fluent if $f(S_c) > 0.5$.
Notice that the correspondence between the English and the translated Chinese sentences allows us to use the same labels from $\mathcal{D}$ to train all the classifiers.

We solve Eq. (\ref{eq:obj}) using stochastic gradient descent with Adam \cite{iclr2014-adam} on batches of size 64. We empirically set the initial learning rate $\eta=0.0001$, decay weights $\beta_1=0.9$, $\beta_2=0.9$ and small constant $\epsilon=10^{-6}$ for Adam. We apply dropout to output of the word embedding layer and LSTM to mitigate model overfitting. The size of the word embeddings and the size of LSTM are both set to be 512. We employ BOSON and a Stanford parser \cite{sf-parser} to acquire Chinese and English POS tags, respectively.

\subsection{Model for Image Captioning}

For the Chinese caption generation model, we follow a popular CNN + LSTM approach developed by Vinyals \etal \cite{google-show-tell}.
More formally, for a given image $I$, we aim to automatically predict a Chinese sentence $S=(w_1,w_2,...,w_n)$ that describes in brief the visual content of the image.
A probabilistic model is used to estimate the posterior probability of a specific sequence of words given the image.
Given $\theta$ as the model parameters, the probability is expressed as $p(S|I;\theta)$.
Applying the chain rule together with log probability for the ease of computation,
we have 
\begin{equation} \label{eq:nic-logprob}
\log p(S|I;\theta) =\sum_{t=1}^{n+1} \log p(w_t|I,w_0,\ldots,w_{t-1};\theta), 
\end{equation}
where $w_0=w_{n+1}=\mbox{START/END}$ is a special token indicating the beginning or the end of the sentence.
Consequently, the image will be annotated with the sentence that yields the maximal posterior probability.

Conditional probabilities in Eq. (\ref{eq:nic-logprob}) are estimated by the LSTM network in an iterative manner.
The LSTM network maintains a cell vector $c$ and a hidden state vector $h$ to adaptively memorize the information fed to it. As shown in Fig. \ref{fig:framework}, the recurrent connections of LSTM carry on previous context. 
In the training stage, pairs of image and translated Chinese sentence are fed to the model. 
At the very beginning, the embedding vector of an image, $x_{-1}$, obtained by applying an affine transformation on its visual representation $CNN(I)$, is fed to the network to initialize the two memory vectors. The word sequence $(w_0,\ldots ,w_n)$, after applying a linear transformation on the word embedding vectors, is iteratively fed to the LSTM.
In the $t$-th iteration, new probabilities $p_t$ over all the candidate words are re-estimated given the current context. 
To express the above process in a more formal way, we write 
\begin{align} 
  x_{-1}&:= W_v \cdot CNN(I),\\
  x_t&:= W_s \cdot \mathbf{w}_t,~~~t=0,1,\ldots,\\
  p_0,c_0, h_0 & \leftarrow \mbox{LSTM}(x_{-1},\mathbf{0},\mathbf{0}),\\
  p_{t+1}, c_{t+1}, h_{t+1} & \leftarrow \mbox{LSTM}(x_t,c_t,h_t).
\end{align}
%
The parameter set $\theta$ consists of $W_v$, $W_s$, and parameters w.r.t. affine transformations inside LSTM.

The loss is the sum of the negative log likelihoods of the next correct word at each step. 
We use SGD with mini-batches of $m$ image-sentence pairs. Given training samples $\{(I_i,S_i)|i=1,...,m\}$ in a batch, the loss is formulated as follows:
$$ bLoss = - \frac{1}{m} \sum_{i=1}^{m} \log p(S_i|I_i;\theta)  $$

In the inference period, after feeding the image embedding vector, the softmax layer after the LSTM produces a probability distribution over all words. The word with the maximum probability is picked up, and fed to LSTM in the next iteration. Following \cite{google-show-tell,cvpr2015-neuraltalk}, per iteration we apply beam search to maintain the $k$ best candidate sentences, with a beam of size 5. The iteration stops once a special END token is selected.

To extract image representations, we use a pre-trained ResNet-152 \cite{resnet2016} which achieved state-of-the-art results for image classification and detection in both ImageNet and COCO competitions. The image feature is extracted as a 2048-dimensional vector from the pool5 layer after ReLU. We conduct l2 normalization on the extracted features since it leads to better results according to our preliminary experiments. 
The dimension of image and word embeddings, and the hidden size of LSTM are all set to be 512. We replace words that occurring less than five times in the training set with a special `UNK' token.
We set the initial learning rate $\eta=0.001$, decaying every ten epochs with a decay weight of 0.999.


\subsection{Fluency-Guided Training}

Having the sentence fluency classifier and the image captioning model introduced, we are now ready to discuss how to guide the training process in light of the estimated fluency and consequently generate better-formed Chinese captions.
While the question is new, if we view fluency as a measure of the importance of the individual training samples, we see some conceptual resemblance to a machine learning scenario where some samples are more important than others. 
A typical case is learning from a data set with highly unbalanced classes,
where one might consider down-sampling classes in majority, over-sampling classes in minority or re-weighting samples \cite{weiss2007cost,ijcai2001-csl}. In our context, fluent sentences are in short supply relatively. 
Inspired by such a connection, we propose three strategies for fluency-guided training,

\textbf{Strategy I: Fluency only}.
This strategy preserves only sentences classified as fluent for training the captioning model.
Models derived from such cleaned dataset tend to generate more fluent captions.
Nonetheless, this benefit is obtained at the risk of learning from insufficient data.
As aforementioned, translated sentences with low fluency can still contain correct keywords which can provide connections between the visual representation and the language model. 
To overcome the downside of the first strategy, the following two strategies are introduced.

\textbf{Strategy II: Rejection sampling}. 
We introduce a sampling-based strategy that allows the sentences classified as not fluent to be used for training with a certain chance, besides preserving all sentences classified as fluent. 
Naturally this chance shall be proportional to the sentences' probability of being fluent. 
As $f(S_c)$ is a classifier output, directly sampling w.r.t. $f(S_c)$ is hard. 
We thus leverage rejection sampling, a type of Monte Carlo method developed for handling such difficulties. 
For a sentence having $f(S_c)<0.5$, a number $u$ is randomly drawn from the uniform distribution $U(0,0.5)$. 
The sentence will be included in the current mini-batch if $f(S_c) > u$, and rejected otherwise.

\textbf{Strategy III: Weighted loss}.
This strategy makes full use of the translated sentences by cost-sensitive learning \cite{ijcai2001-csl}. 
In particular, we multiply the fluency score $f(S_i)$ to a training sample's loss as a penalty weight when calculating the loss in every mini-batch. In particular, the weighted loss for a mini-batch is computed as
\begin{equation}
bLoss_{weighted} = - \frac{1}{m} \sum_{i=1}^{m} \mu_i \cdot \log p(S_i|I_i;\theta),
\end{equation}
where $\mu_i=1$ if $f(S_c)>0.5$, \ie classified as fluent, otherwise $\mu_i=f(S_c)$.

In what follows we will evaluate the viability of the three fluency-guided training strategies.

\section{Experiments} \label{sec:eval}

The main purpose of our experiments is to verify if a cross-lingual captioning model trained by fluency-guided learning can generate Chinese captions that are more fluent, meanwhile maintaining the level of relevance when compared to learning from the complete set of machine-translated sentences. We term this baseline as `Without fluency'. 
As sentence fluency estimation is a prerequisite for fluency-guided learning, we first evaluate this component.

\subsection{Sentence Fluency Estimation}

\textbf{Setup}. 
In order to train the four-way sentence fluency classifier, a number of paired bilingual sentences labeled as fluent / not fluent are a prerequisite.
We aim to select a representative and diverse set of sentences for manual verification, meanwhile keeping the manual annotation affordable.
To this end we sample at random 2k and 6k English sentences from Flickr8k~\cite{flickr8k} and MSCOCO~\cite{mscoco} respectively. The 8k sentences were automatically translated into the same amount of Chinese sentences by the Baidu translation API.
Manual verification was performed by eight students (all native Chinese speakers) in our lab.
In particular, each Chinese sentence was separately presented to two annotators, asking them to grade the sentence as \textit{fluent}, \textit{not fluent}, or \textit{difficult to tell}. 
A sentence is considered fluent if it does not contain obvious grammatical errors and is in line with language habits of Chinese. 
Sentences receiving inconsistent grades or graded as \textit{difficult to tell} were ignored.
This resulted in 6,593 labeled sentences in total.
They are then randomly split into three folds, \ie 4,593 / 1,000 / 1,000 for training / validation / test, as summarized in Table \ref{tab:data_statistics}. 
The fact that less than 30\% of the translated sentences are considered fluent indicates much room for further improvement for the current machine translation system. It also shows the necessity of fluency-guided learning when deriving cross-lingual image captioning models from machine-translated corpus.

\begin{table} [b!]
\renewcommand{\arraystretch}{1.2}
\caption{Datasets for sentence fluency estimation.
The relatively low rate of fluency (less than 30\%) in machine-translated sentences indicates the importance of fluency-guided learning for cross-lingual image captioning. }
\label{tab:data_statistics}
\centering
 \scalebox{0.9}{
\begin{tabular}{@{}l rrr@{}}
\toprule
   & \textbf{training} & \textbf{validation} & \textbf{test}  \\
\cmidrule{1-4}
\# fluent           & 1,240  & 291 & 294 \\
\# not fluent       & 3,353  & 709 & 706 \\
\bottomrule
\end{tabular}
 }
\end{table}

\textbf{Baselines}. In order to obtain a more comprehensive picture, we consider two baselines. One is random guess. The other is to predict fluency in terms of sentence length. This is based on our observation that longer sentences are more difficult to be translated. In particular, a sentence, let it be $S_e$ or $S_c$, is classified as fluent if its length is less than the average length of the fluent sentences in the training set.

\textbf{Results}.
Table \ref{tab:class_perf} shows the performance of different models for sentence fluency classification on the test set.
The proposed four-way LSTM achieves the highest precision at the cost of recall. 
This is desirable as sentences incorrectly classified as not fluent still have a chance to get back in the subsequent fluency-guided learning stage.
Some qualitative results are provided in Table \ref{tab:fluency-results}.

\begin{table} [tb!]
\renewcommand{\arraystretch}{1.2}
\caption{Performance of varied models for sentence fluency classification.
The four-way LSTM achieves the highest precision, at the cost of recall.}
\label{tab:class_perf}
\centering
 \scalebox{0.99}{
\begin{tabular}{@{}l rrrr@{}}
\toprule
\textbf{Model}  && \textbf{Recall} & \textbf{Precision}  \\
\cmidrule{1-1} \cmidrule{3-4}
random guess            && 50.0  & 29.4  \\ 
Length of $S_e$         && 48.3  & 39.8  \\ 
Length of $S_c$         && 49.3  & 45.3  \\ [3pt]
LSTM(English words)      && 37.1  &   58.0    \\
LSTM(English POS tags)        && 21.1  &   58.0    \\
LSTM(Chinese words)      && \textbf{50.3}  &   61.7    \\
LSTM(Chinese POS tags)    && 44.9  &   62.6    \\
Four-way LSTM classifier  && 34.0  &   \textbf{80.0}    \\
\bottomrule
\end{tabular}
 }
\end{table}



\begin{table} [tb!]
\renewcommand{\arraystretch}{1.4}
\caption{Examples of sentence fluency estimation by our four-way LSTM classifier. 
For those sentences receiving lower fluency scores, 
while keywords in the English sentences are correctly translated, 
their conjunction is inappropriate, making the translated sentences unreadable.
}
\label{tab:fluency-results}
\centering
 \scalebox{0.85}{
\begin{tabular}{@{}p{4cm} p{4.2cm} r@{}}
\toprule
\textbf{English sentence} $S_e$ & \textbf{Machine translated sentence} $S_c$ & $f(S_c)$  \\
\cmidrule{1-3}
The two large elephants are standing in the grass                 & 两 只 大象 正 站 在 草地 上                       & 0.803  \\   
The young man in the blue shirt is playing tennis                 & 穿 蓝色 衬衫 的 年轻人 正在 打 网球                & 0.624  \\
A male tennis player in action before a crowd                     & 一 名 男子 网球 运动员 在 人群 前 行动             & 0.424  \\ 
A couple of people on a \mbox{motorcycle} posing for a picture    & 一 对 夫妇 的 摩托车 冒充 一个 图片                & 0.219 \\
Many stuffed teddy bears are set next to one another              & 许多 毛绒 玩具 熊 被 设置 在 另 一个               & 0.158 \\
A group of people riding skis in their bathing suits              & 一 群 人 在 他们 的 沐浴 骑 滑雪 服                & 0.117 \\
A sports arena under a dome with snow on it                       & 一个 体育馆 下 一个 圆顶 下 的 雪 在 它             & 0.060 \\
\bottomrule
\end{tabular}
 }
\end{table}

\subsection{Image Caption Generation}

\textbf{Setup}. 
While we target at learning from machine-translated corpus, manually written sentences are needed to evaluate the effectiveness of the proposed framework. To the best of our knowledge, Flickr8k-cn \cite{icmr2016} is the only public dataset suited for this purpose.
Each test image in Flickr8k-cn is associated with five Chinese sentences, obtained by manually translating the corresponding five English sentences from Flickr8k \cite{flickr8k}.
In addition to Flickr8k-cn,  we construct another test set by extending Flickr30k \cite{flickr30k} to a bilingual version.
For each image in the Flickr30k training / validation sets, we employ Baidu translation to automatically translate its sentences from English to Chinese. The sentences associated with the test images are manually translated. 
Similar to \cite{icmr2016}, we hire five Chinese students who are fluent in English (passing the national College English Test 6).
Notice that an English word might have multiple translations, \eg football can be translated into `足球'(soccer) and `橄榄球'(American football). 
For disambiguation, translators were shown an English sentence together with the image.
For the sake of clarity, we use Flickr30k-cn to denote the bilingual version of Flickr30k.
Besides the translation of English captions, Flickr8k-cn also contains independent manually written Chinese captions. 
Main statistics of Flickr8k-cn and Flickr30k-cn are given in Table \ref{tab:caption_data}.

\begin{table} [tb!]
\renewcommand{\arraystretch}{1.2}
\caption{Two datasets used in our image captioning experiments.
Besides Flickr8k-cn \cite{icmr2016}, we construct Flickr30k-cn, a bilingual version of Flickr30k \cite{flickr30k} obtained by English-to-Chinese machine translation of its train / val sets and human translation of its test set.}
\label{tab:caption_data}
\centering
 \scalebox{0.82}{
\begin{tabular}{@{}l l r r r rr r r@{}}
\toprule
&& \multicolumn{3}{c}{\textbf{Flickr8k-cn}~\cite{icmr2016}} && \multicolumn{3}{c}{\textbf{Flickr30k-cn} (this work)} \\
\cmidrule{3-5} \cmidrule{7-9}
&& train & val & test && train & val & test \\
\cmidrule{1-9}
Images  && 6,000  & 1,000 & 1,000 && 29,783 & 1,000 &1,000\\
\cmidrule{1-1} \cmidrule{3-5} \cmidrule{7-9}
\specialcell{Machine-translated\\Chinese sentences}  && 30,000  & 5,000 & -- && 148,915 & 5,000 & --\\
\cmidrule{1-1} \cmidrule{3-5} \cmidrule{7-9}
\specialcell{Human-translated\\Chinese sentences}  && --  & -- & 5,000 && -- & -- &5,000\\
\cmidrule{1-1} \cmidrule{3-5} \cmidrule{7-9}
\specialcell{Human-annotated\\Chinese sentences}  && 30,000  & 5,000 & 5,000 && -- & -- & --\\
\bottomrule
\end{tabular}
 }
\end{table}




\textbf{Baselines}.
To verify the effectiveness of our fluency-guided approach, we compare with the following three alternatives: 
\begin{enumerate}
\item `Late translation'~\cite{icmr2016}, which generates Chinese captions by automatically translating the output of an English captioning model. 
\item `Late translation rerank', which reranks the top 5 sentences generated by `Late translation' according to their estimated fluency scores in descending order. 
\item `Without fluency', which learns from the full set of machine-translated sentences.
\end{enumerate}
Furthermore, to understand the performance gap between the proposed approach and the method directly using manually written Chinese captions, we train a Chinese model using Flickr8k-cn~\cite{icmr2016}, the only dataset that provides manually written Chinese captions for training. We term this model `Manual Flickr8k-cn'. 



\textbf{Automated evaluation}.
We adopt performance metrics widely used in the literature, \ie BLEU-4, ROUGE-L, and CIDEr. 
The only exception is METEOR \cite{denkowski:lavie:meteor-wmt:2014}, which is inapplicable for evaluating Chinese sentences due to the lack of a structured thesaurus such as WordNet in Chinese. 
BLEU is originally designed for automatic machine translation where they compute the geometric mean of n-gram based precision for the candidate sentence with respect to the references and adds a brevity-penalty to discourage overly short sentences~\cite{bleu}. 
ROUGE is an evaluation metric based on F-measure of longest common sub-sequence~\cite{rouge}.
CIDEr is a metric developed specifically for evaluating image captioning~\cite{cider}. 
It performs a Term Frequency Inverse Document Frequency (TF-IDF) weighting for each n-gram to give less-informative n-grams lower weight. 
The CIDEr score is computed using average cosine similarity between the candidate sentence and the reference sentences.
We use the coco-evaluation code\footnote{\url{https://github.com/tylin/coco-caption}} to compute the three metrics, using human translated captions as ground truth.

Performance on the automatically computed metrics of different approaches is presented in Table \ref{tab:result_auto}. 
The reranking strategy improves over `Late translation' showing the benefit of fluency modeling. Nevertheless, both `Late translation' and `Late translation rerank' perform worse than the `Without fluency' run.
Fluency-only is inferior to other proposed approaches as this model is trained on much less amounts of data, more concretely, 2,350 sentences in Flickr8k and 15,100 sentences in Flickr30k that are predicted to be fluent. 
Both rejection sampling and weighted loss are on par with the `Without fluency' run,
showing the effectiveness of the two strategies for preserving relevant information.


\begin{table} [tb!]
\renewcommand{\arraystretch}{1.2}
\caption{Automated evaluation of six approaches to cross-lingual image captioning. 
Rejection sampling and weighted loss are comparable to `Without fluency' which learns from the full set of machine-translated sentences.}
\label{tab:result_auto}
\centering
 \scalebox{0.83}{
\begin{tabular}{@{}lrrrrrrr@{}}
\toprule
\multirow{2}{*}{\textbf{Approach}} & \multicolumn{3}{c}{\textbf{Flickr8k-cn}} && \multicolumn{3}{c}{\textbf{Flickr30k-cn}} \\ 
\cmidrule(l){2-4}\cmidrule{6-8} 
                          & B-4     & ROUGE    & CIDEr    && B-4     & ROUGE     & CIDEr    \\ 
\midrule
Late translation          & 17.3      & 39.3     & 33.7     && 15.3      & 38.5      & 27.1     \\
Late translation rerank   & 17.5      & 40.2     & 34.2     && 14.3      & 38.5      & 27.5    \\
Without fluency         &\textbf{24.1} &\textbf{45.9}&\textbf{47.6}&& 17.8      &\textbf{40.8}& 32.5     \\ [3pt]
\emph{Fluency-only}              & 20.7      & 41.1     & 35.2     && 14.5      & 35.9      & 25.1     \\
\emph{Rejection sampling}        & 23.9      & 45.3     & 46.6     && 18.2      & 40.5      & 32.9     \\
\emph{Weighted loss}             & 24.0      & 45.0     & 46.3     && \textbf{18.3}& 40.2  &\textbf{33.0} \\ 

\bottomrule

\end{tabular}
 }
\end{table}



\textbf{Human evaluation}. Although BLEU \cite{bleu} is designed to account for fluency, it has been criticized
in the context of machine translation for being loosely approximate human judgments~\cite{callison2006re}. In particular, the n-gram based measure is insufficient to guarantee the overall fluency of a generated sentence.
We therefore perform a human evaluation as follows. Given a test image, sentences generated by distinct approaches are shown together to a subject, who is to rate the sentences using a Likert scale of 1 to 5 (higher is better) in two aspects, namely relevance and fluency. While rating is inevitably subjective, putting the sentences together helps the subject provide more comparable scores. Eight persons in our lab including paper authors participate the evaluation. Notice that to avoid bias, sentences are always randomly shuffled before presenting to the subjects. To reduce the workload, the evaluation is performed on a random subset of 100 images for each test set, and each image is rated by two distinct subjects. Average scores are reported.

As shown in Table \ref{tab:result_human}, 
the reranking strategy results in more fluent captions compared to the `Late translation' approach, improving fluency from 4.34 to 4.41 on Flickr8k-cn and from 4.60 to 4.75 on Flickr30k-cn, showing the effectiveness of the proposed LSTM classifier for sentence fluency estimation. 

On both test sets, the three proposed strategies improve the fluency of the generated captions compared to the baselines. Though receiving high fluency rate on Flickr8k-cn, the fluency-only model still suffers from lower relevance. The user study suggests that rejection sampling outperforms weighted loss in terms of both relevance and fluency.
In addition, we find that for rejection sampling, the average number of mini-batches in each training epoch is 75 on Flickr8k and 616 on Flickr30k, which is less than half of the number of mini-batches for weighted loss.
Compared to `Late translation rerank', rejection sampling performs better in describing images, suggesting that both relevance and fluency have to be taken into account for cross-lingual image captioning.

Model trained on manual annotation performs better than fluency-guided learning on Flickr8k-cn, improving relevance from 3.27 to 3.32 and fluency from 4.66 to 4.79. 
However, the model is less effective when tested on Flickr30k-cn, with relevance decreased from 3.20 to 2.83 and fluency from 4.76 to 4.12. 
Learning from many translated text guided by fluency results in cross-lingual models with better generalization ability.


\begin{table} [tb!]
\renewcommand{\arraystretch}{1.1}
\caption{Human evaluation of seven approaches to cross-lingual image captioning.
Rejection sampling achieves the best balance between relevance and fluency, without the need of manual written Chinese captions.}
\label{tab:result_human}
\centering
 \scalebox{0.85}{
\begin{tabular}{@{}lrrrrr@{}}
\toprule
\multirow{2}{*}{\textbf{Approach}} & \multicolumn{2}{c}{\textbf{Flickr8k-cn}} && \multicolumn{2}{c}{\textbf{Flickr30k-cn}} \\ 
\cmidrule{2-3} \cmidrule{5-6} 
                       & Relevance      & Fluency        && Relevance       & Fluency        \\ 
\midrule
Late translation       & 2.91 $\pm 1.11$ & 4.34 $\pm 1.11$ && 3.00 $\pm 1.01$& 4.60 $\pm 0.65$\\
Late translation rerank       & 3.04 $\pm 1.14$ & 4.41 $\pm 0.92$ && 3.14 $\pm 1.01$& 4.75 $\pm 0.44$\\

Without fluency      & 3.18 $\pm 1.09$ & 4.12 $\pm 1.09$ && 3.06 $\pm 0.93$& 4.21 $\pm 1.13$\\ [3pt]

\emph{Fluency-only}    & 2.67 $\pm 1.06$ & 4.76 $\pm 0.43$&&2.58 $\pm 0.98$& 4.74 $\pm 0.42$ \\

\emph{Rejection sampling}& 3.27 $\pm 1.04$  & 4.66 $\pm 0.59$&& \textbf{3.20} $\pm 0.96$& \textbf{4.76} $\pm 0.48$\\

\emph{Weighted loss}        & 3.23 $\pm 1.11$ & 4.66 $\pm 0.51$&& 2.96 $\pm 1.02$& 4.68 $\pm 0.52$\\[3pt]
Manual Flickr8k-cn    & \textbf{3.32} $\pm 0.94$ & \textbf{4.79} $\pm 0.38$ && 2.83 $\pm 1.22$& 4.12 $\pm 1.41$\\ 
\bottomrule
\end{tabular}
 }
\end{table}

For a more intuitive understanding, some qualitative results are shown with human evaluation in Table \ref{tab:caption-results}.

\begin{table*} [tb!]
\renewcommand{\arraystretch}{1.1}
\caption{\textbf{Bilingual captions generated by eight approaches}.
1) English: An English captioning model.
2) Late translation: Machine translation of the previous English sentence.
3) Late translation rerank: Reranking the output of `Late translation' by estimated fluency scores.
4) Without fluency: A Chinese captioning model trained on machine-translated sentences without considering sentence fluency.
5) Fluency-only: A Chinese captioning model trained on machine-translated sentences classified as fluent.
6) Rejection sampling: Favor training sentences with larger fluency scores.
7) Weighted loss: Penalize training sentences in terms of their estimated fluency scores.
8) Manual Flickr8k-cn: A Chinese captioning model trained on manually written Chinese captions.
Human evaluation is also presented as a tuple (relevance, fluency) after each Chinese sentence.
}
\label{tab:caption-results}
\centering
   \scalebox{.85}{
    \begin{tabular}{@{}| l | ll | ll |@{}}
    \toprule
 &\raisebox{-0.5\totalheight}{\includegraphics[width=3.8cm,height=3.8cm]{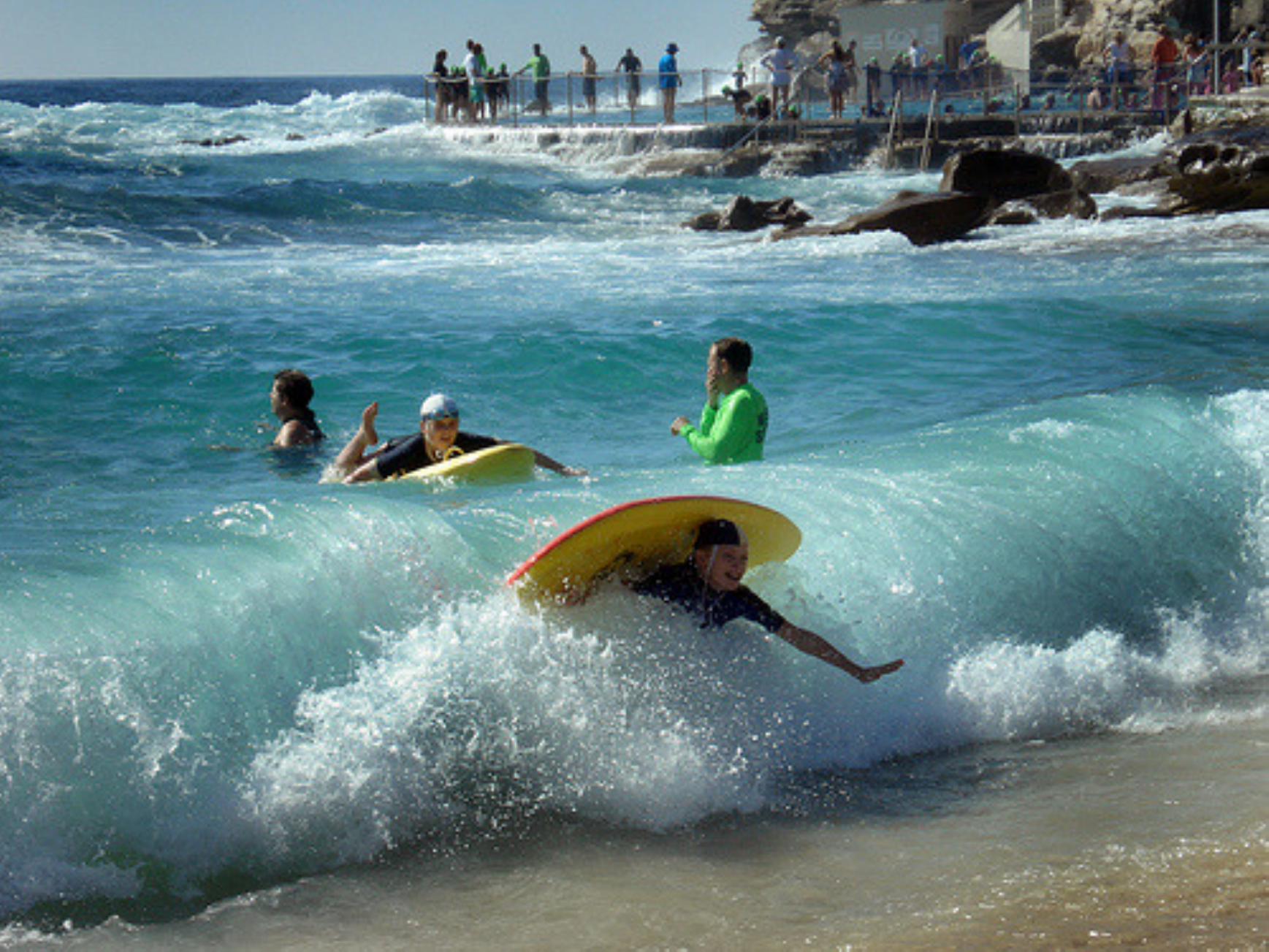}} &&  \raisebox{-0.5\totalheight}{\includegraphics[width=3.8cm,height=3.8cm]{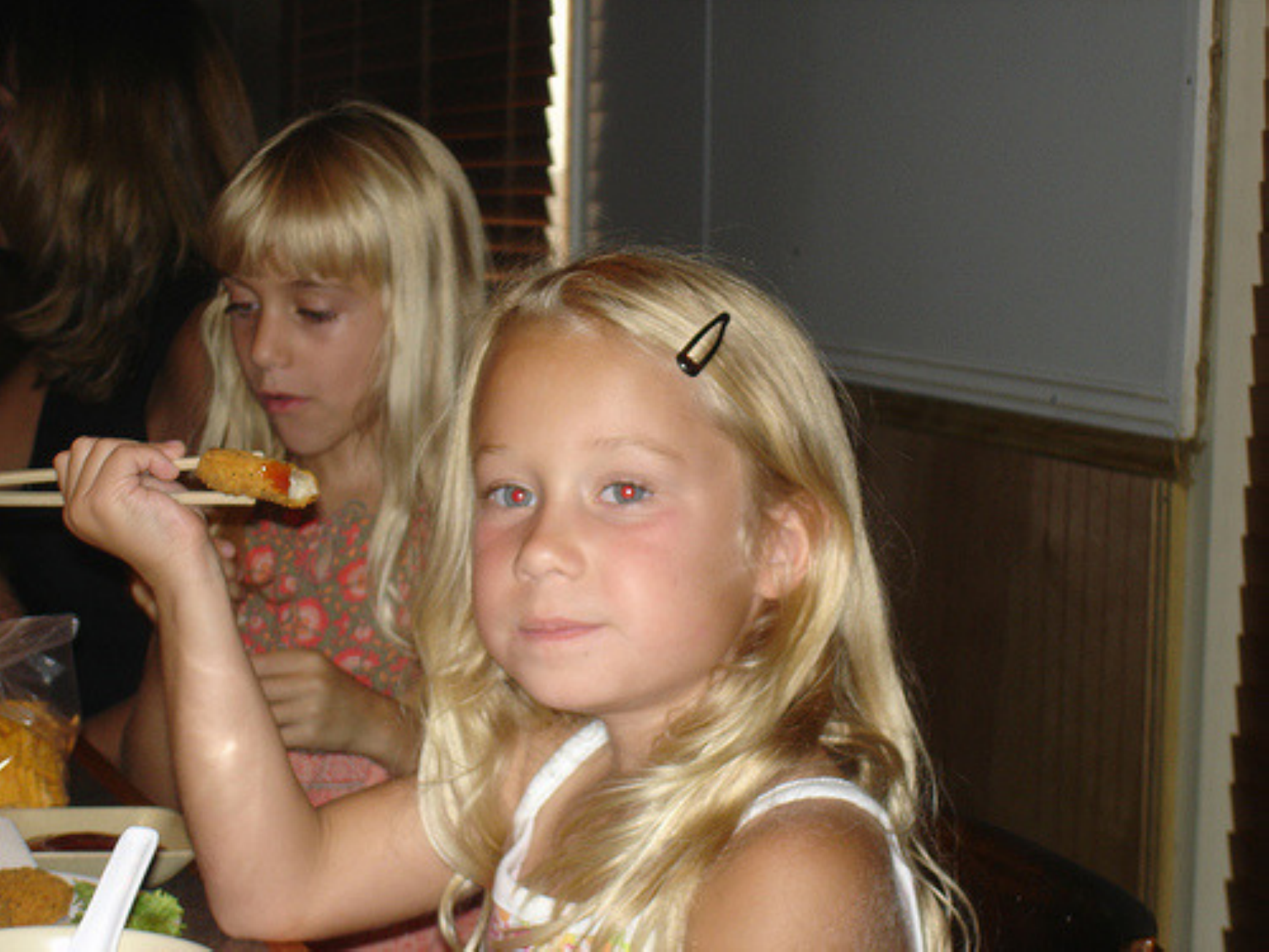}}& \\
  English & a surfer rides a wave && a little girl in a pink shirt is eating&\\
  Late translation & 冲浪者 骑 波 &(3.5, 3.0)&一个 穿 着 粉红色 衬衫 的 小 女孩 正在 吃饭 & (3.5, 4.5)\\
  Late translation rerank & 冲浪者 骑 浪 &(4.0, 3.0)& 一个 穿 粉红色 衬衫 的 小 女孩 在 吃 东西  & (3.5, 4.5) \\
  Without fluency &冲浪者 骑 波 &(3.5, 3.0) &一个 年轻 的 女孩 在 一个 粉红色 的 衬衫 拿 着 一个 粉红色 的 &(2.0, 1.5)  \\
\emph{Fluency-only} &一个 人 在 水 里 游泳 &(3.5, \textbf{5.0}) &  一个 小 女孩 抱 着 一个 婴儿 &(2.0, \textbf{5.0}) \\
\emph{Rejection sampling}&一个 人 在 海洋 里 冲浪 &(\textbf{4.5}, \textbf{5.0}) & 一个 年轻 的 金发 女孩 正在 吃 东西 & (\textbf{4.5}, 4.5)  \\
\emph{Weighted loss} &一个 人 在 海洋 里 冲浪 &(\textbf{4.5}, \textbf{5.0}) & 两 只 小 女孩 在 一 张 桌子 上 吃 东西& (\textbf{4.5}, 4.5)  \\
 Manual Flickr8k-cn & 一个 男人 在 海上 冲浪 & (\textbf{4.5}, \textbf{5.0}) & 一个 女人 抱 着 一个 小孩 & (2.0, \textbf{5.0}) \\
 \midrule
 &\raisebox{-0.5\totalheight}{\includegraphics[width=3.8cm,height=3.8cm]{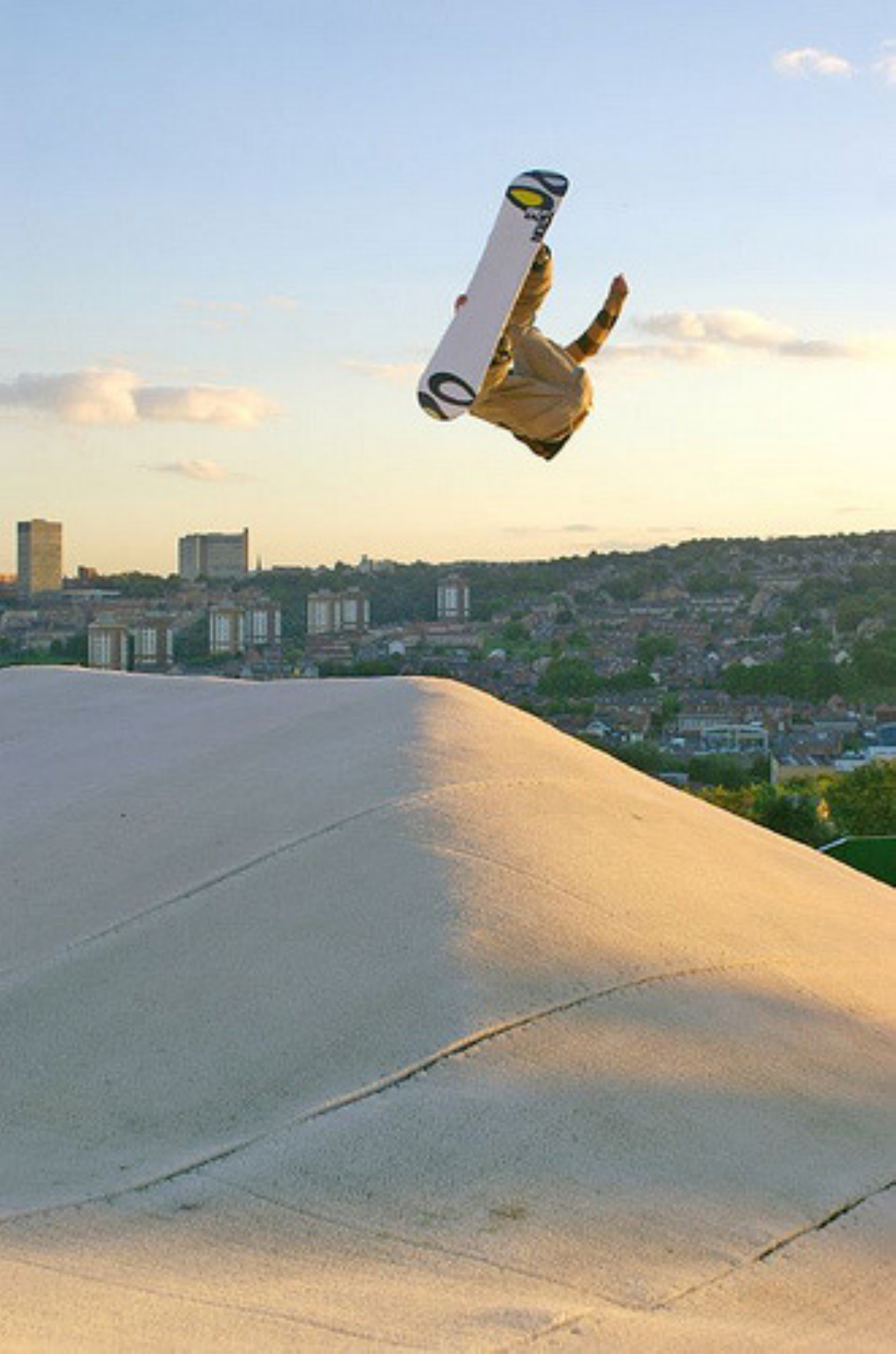}} &&  \raisebox{-0.5\totalheight}{\includegraphics[width=3.8cm,height=3.8cm]{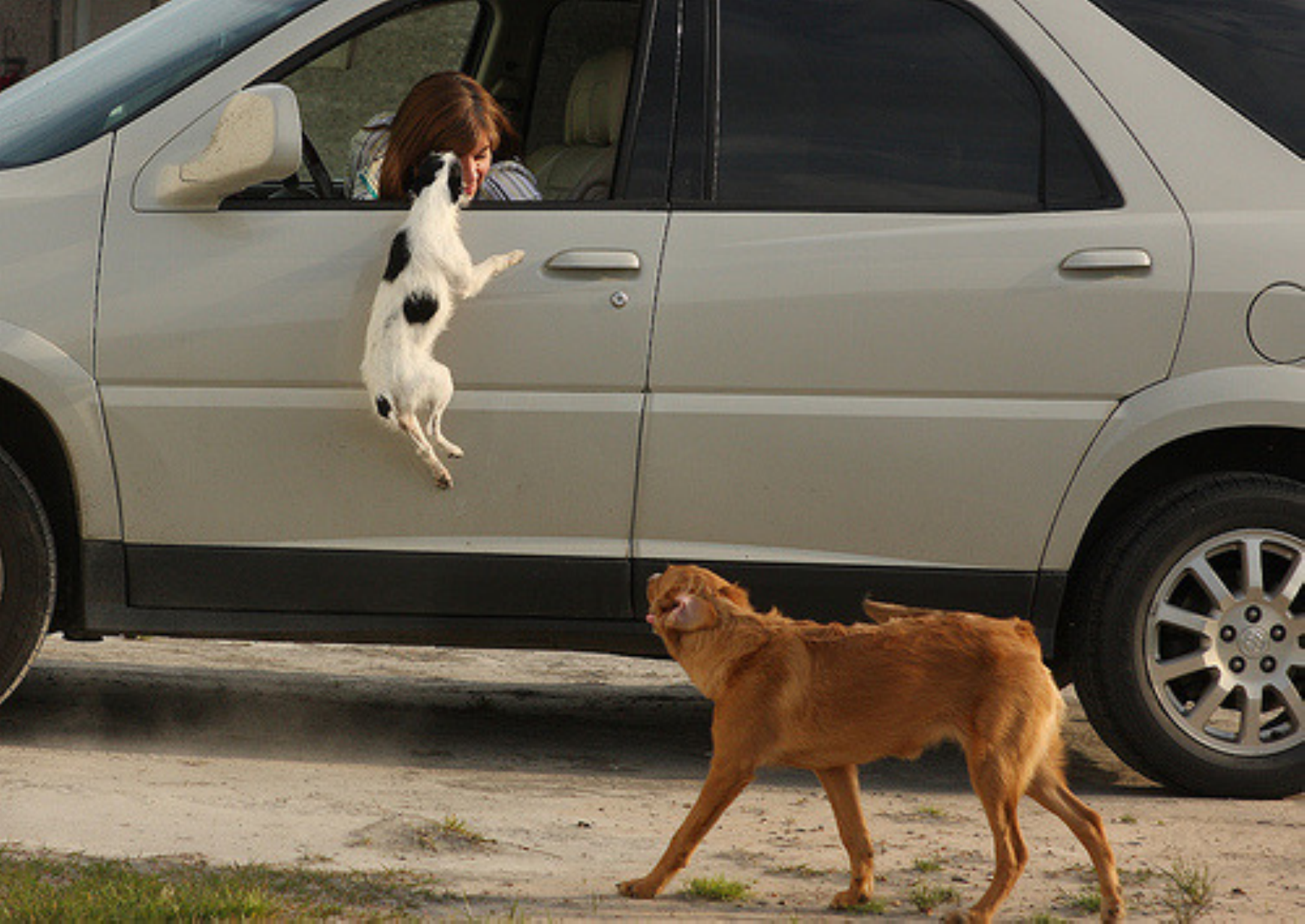}}& \\
  English & a skateboarder is doing a jump && two dogs play in a yard&\\
  Late translation & 一个 滑板 做 跳 &(2.5, 3.0)& 两 只 狗 在 院子 里 玩耍 &(3.0, \textbf{5.0})\\
  Late translation rerank & 一个 滑板 做 跳 &(2.5, 3.0) & 两 只 狗 在 院子 里 玩耍 & (3.0, \textbf{5.0}) \\
  Without fluency &  一个 人 在 空中 跳跃 &(3.0, \textbf{5.0}) & 一 只 棕色 的 狗 和 一 只 白色 的 狗 在 一 条 黑色 的 &(3.0, 3.5)\\
\emph{Fluency-only} & 一个 人 爬 上 了 一 座 岩 石墙 &(2.0, \textbf{5.0})& 一 只 棕色 的 狗 跳 过 了 一个 障碍 &(2.0, \textbf{5.0})\\
\emph{Rejection sampling}& 一个 人 在 空中 跳跃 &(3.0, \textbf{5.0})& 一 只 白色 的 狗 和 一 只 棕色 的 狗 在 街上 &(\textbf{3.5}, \textbf{5.0})\\
\emph{Weighted loss} & 一个 人 在 空中 跳跃 &(3.0, \textbf{5.0})& 一 只 狗 在 沙滩 上 玩 球 &(2.0, 4.5)\\
 Manual Flickr8k-cn &  一个 人 在 玩 滑板& (\textbf{4.0}, \textbf{5.0})  & 两 只 狗&(2.5, \textbf{5.0})\\
 \midrule
 &\raisebox{-0.5\totalheight}{\includegraphics[width=3.8cm,height=3.8cm]{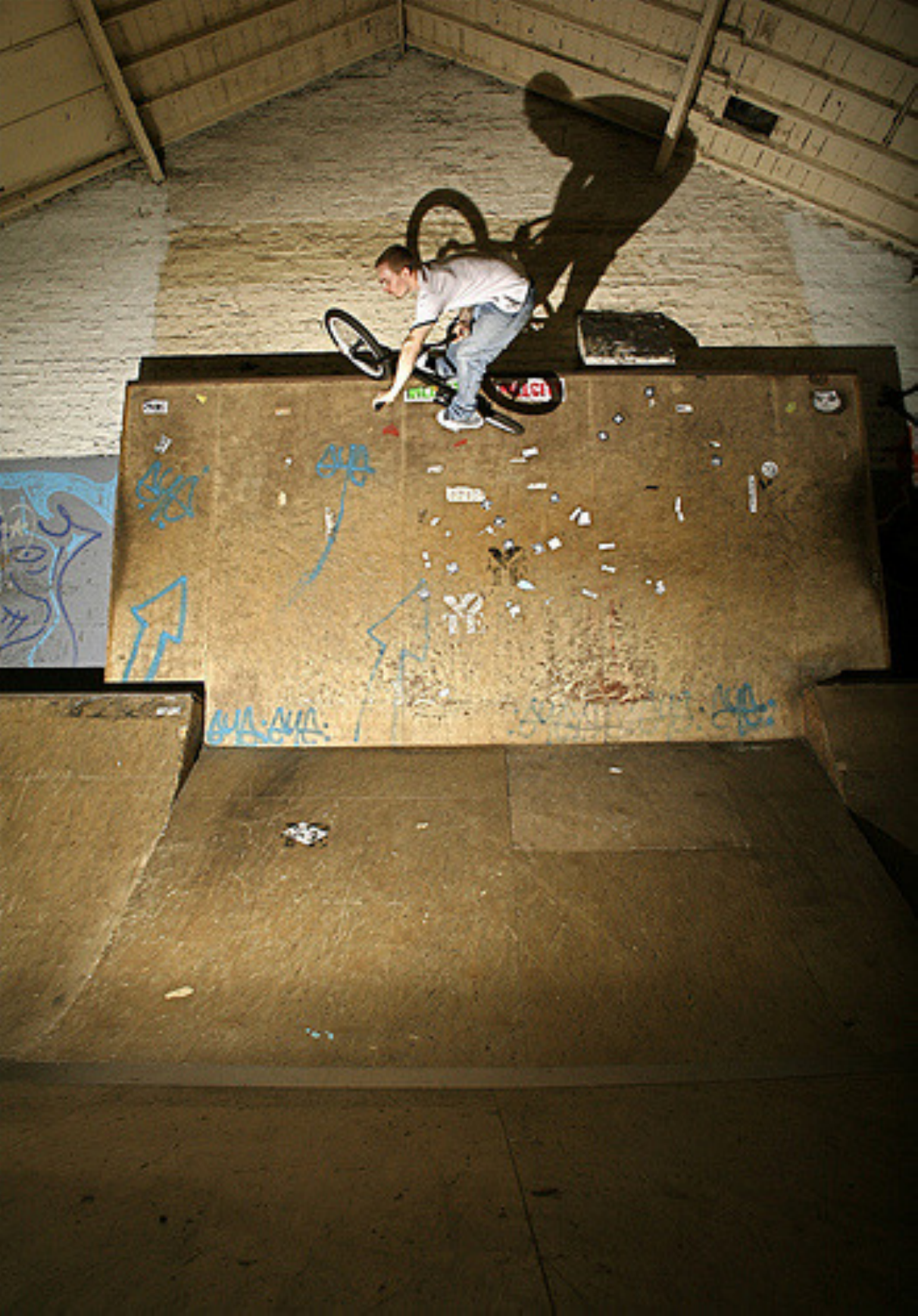}} &&  \raisebox{-0.5\totalheight}{\includegraphics[width=3.8cm,height=3.8cm]{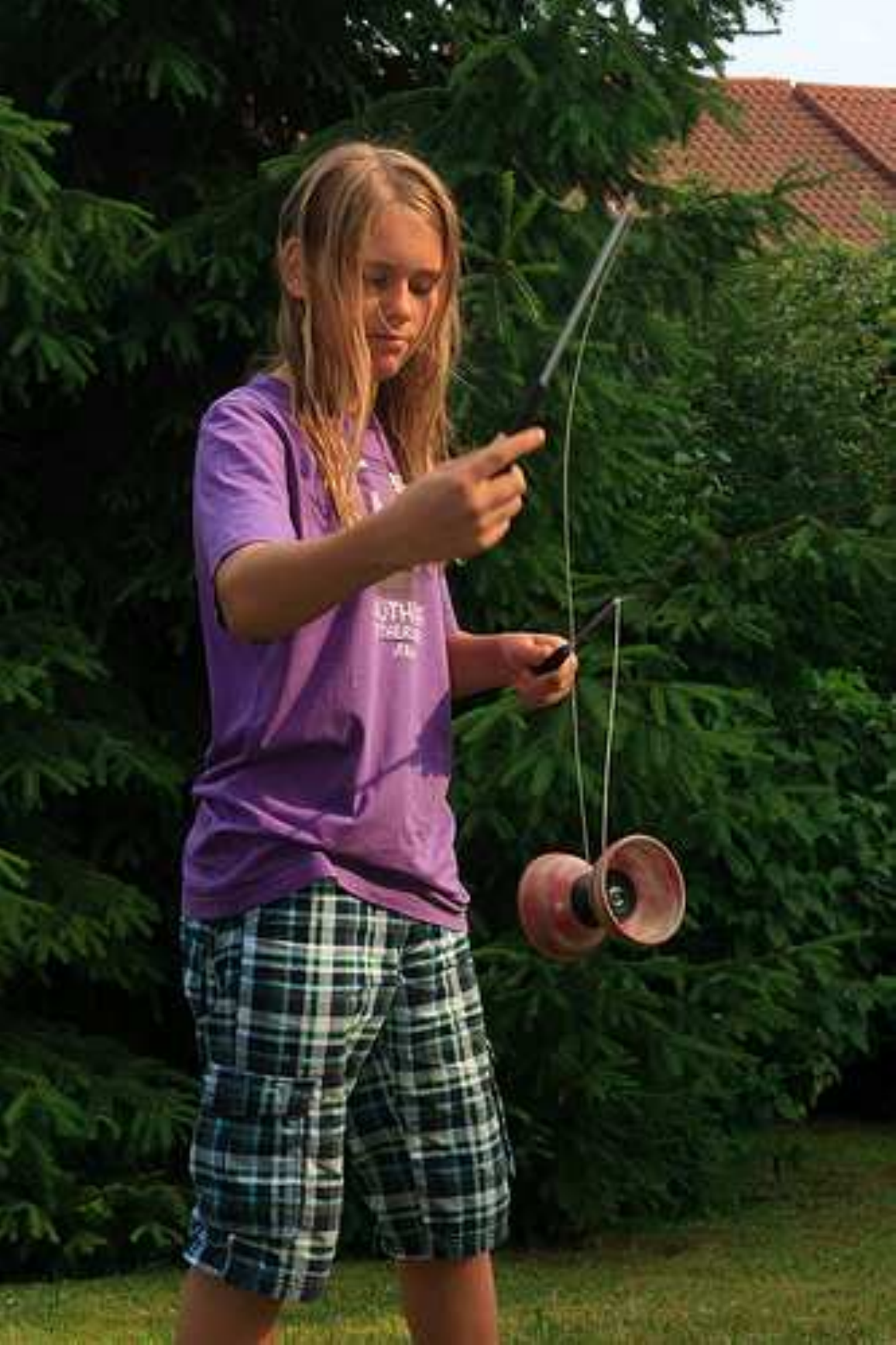}}& \\

  English & a skateboarder does a trick on a ramp && a young girl in a pink shirt is playing a game &\\
  Late translation &一个 滑板 在 斜坡 的 把戏 &(3.0, 4.0) &  一个 穿 着 粉红色 衬衫 的 年轻 女孩 正在 玩 游戏 &(\textbf{3.5}, \textbf{5.0}) \\
  Late translation rerank & 一个 滑板 在 斜坡 的 把戏 &(3.0, 4.0) &  一个 穿 粉红色 衬衫 的 年轻 女孩 正在 玩 游戏& (\textbf{3.5}, \textbf{5.0}) \\
Without fluency & 一个 滑板 跳 下 楼梯 &(2.5, 3.5) & 一个 年轻 的 女孩 穿 着 一 件 红色 的 衬衫
 和 蓝色 的 裤子 是 在 一个 UNK &(2.0, 4.0)\\
\emph{Fluency-only} &一个 人 爬 上 一 块 岩石 &(2.0, \textbf{5.0})  & 一个 小 女孩 正在 玩 一个 游戏 &(\textbf{3.5}, \textbf{5.0})\\
\emph{Rejection sampling} &一个 滑板 跳跃 &(2.5, 3.0) & 一个 穿 着 黄色 衬衫 的 小 女孩 在 玩 玩具 &(\textbf{3.5}, \textbf{5.0})\\
\emph{Weighted loss} &一个 人 爬 上 一 块 岩 石墙 &(2.0, \textbf{5.0}) & 一个 小 女孩 在 外面 玩 泡泡
 &(2.5, \textbf{5.0})\\
 Manual Flickr8k-cn & 一个 男人 在 玩 花样 滑板 & (\textbf{4.0}, \textbf{5.0}) & 一个 小 男孩 在 玩耍& (3.0, \textbf{5.0}) \\

    \bottomrule
    \end{tabular}
    }
\end{table*}

\subsection{Discussion}

While we investigate English-to-Chinese as an instantiation of cross-lingual image captioning, the proposed method can be easily extended to another target language, given the availability of some fluency annotations in that language. 
Notice that compared to manually writing sentences for training images given the associated English captions and their machine translation results,
manual annotation effort for fluency modeling is much less. Labeling fluency just needs a click. By contrast, one has to perform a number of edits on the provided translated caption when the translation is unsatisfactory. According to our experiments, 89\% of the provided translations are reedited by annotators. Consequently, on average it takes around 64 seconds to get a decent Chinese caption, while only 5 seconds to obtain a fluency label. So collecting fluency annotation is more efficient. Moreover, the fluency labels are discrete, allowing us to easily obtain consistent and reliable fluency annotation by majority voting on labels from distinct annotators. Also note that fluency prediction as binary classification is less challenging than caption generation, so less amount of training samples is needed.
In summary, fluency-guided learning allows us to perform cross-lingual image captioning with affordable annotation efforts.

\section{Conclusions} \label{sec:conc}

In this paper, we present an approach to cross-lingual image captioning by utilizing machine translation.
A fluency-guided learning framework is proposed to deal with the lack of fluency in machine-translated sentences.
Experiments on two English-Chinese datasets, \ie Flickr8k-cn and Flickr30-cn, support our conclusions as follows.
Less than 30\% of the translated sentences are considered fluent, indicating much room for further improvement for current machine translation.
Meanwhile, the proposed fluency-guided learning by rejection sampling effectively attacks the challenge.
When measured by BLEU-4, ROUGE and CIDEr which emphasize on predicting relevant terms, 
the proposed approach is on par with the baseline that learns from all the translated sentences.
Human evaluation shows that our approach outperforms the baseline in terms of both relevance and fluency.

Our proposed fluency-guided learning framework takes a substantial step towards practical use of machine translation for cross-lingual image captioning with minimal manual annotation efforts. 
Extending our work to multimedia content analysis and repurposing in a multilingual setting opens up promising avenues for future research. 




\section*{Acknowledgments}

This work was supported by National Science Foundation of China (No. 61672523, 71531012). We thank the anonymous reviewers for their insightful comments. A Titan X Pascal GPU used for this research was donated by the NVIDIA Corporation.



\clearpage\end{CJK*}
\end{document}